\documentclass[preprint]{IEEEtran}
\IEEEoverridecommandlockouts
\usepackage{amsmath,amssymb,amsfonts}
\usepackage{algorithmic}
\usepackage{graphicx}
\usepackage{textcomp}
\usepackage{xcolor}
\usepackage{adjustbox}
\usepackage{inputenc}
\usepackage{array}
\usepackage{mdwmath}
\usepackage{mdwtab}
\usepackage{eqparbox}
\usepackage{url}
\usepackage{lineno}
\usepackage[colorlinks=true, allcolors=blue]{hyperref}
\usepackage{mathtools}
\usepackage{subfig}
\usepackage[numbers]{natbib} 
\usepackage[left=16.9 mm, right=16.9 mm, top=16.9 mm, bottom=16.9 mm]{geometry}
\usepackage[normalem]{ulem}
\usepackage{xcolor}
\usepackage{tikz}
\usepackage[flushleft]{threeparttable}
\usepackage{amsmath}
\DeclareMathOperator*{\mmax}{max}
\DeclareMathOperator*{\mmin}{min}

\usepackage{multirow}
\def\BibTeX{{\rm B\kern-.05em{\sc i\kern-.025em b}\kern-.08em
    T\kern-.1667em\lower.7ex\hbox{E}\kern-.125emX}}
\begin{document}
\bibliographystyle{IEEEtranN}

\title{Optimal Control Policies to Address the Pandemic Health-Economy Dilemma \\
}

\author{
\IEEEauthorblockN{Rohit Salgotra, Amiram Moshaiov}
\IEEEauthorblockA{\textit{School of Mechanical Engineering} \\
\textit{Iby and Aladar Fleishman Faculty of Engineering}\\
Tel Aviv University, Israel \\
Email: rohits@mail.tau.ac.il, moshaiov@tauex.tau.ac.il\newline}
\and 
\IEEEauthorblockN{\newline Thomas Seidelmann, Dominik Fischer, Sanaz Mostaghim}
\IEEEauthorblockA{\textit{Chair of Computational Intelligence} \\
\textit{Faculty of Computer Science}\\
Otto von Guericke University Magdeburg, Germany \\
Email: [thomas.seidelmann,dfischer,sanaz.mostaghim]@ovgu.de}

}

\maketitle
\begin{abstract}
Non-pharmaceutical interventions (NPIs) are effective measures to contain a pandemic. Yet, such control measures commonly have a negative effect on the economy. Here, we propose a macro-level approach to support resolving this Health-Economy Dilemma (HED). First, an extension to the well-known SEIR model is suggested which includes an economy model.  Second, a bi-objective optimization problem is defined to study optimal control policies in view of the HED problem.  Next, several multi-objective evolutionary algorithms are applied to perform a study on the health-economy performance trade-offs that are inherent to the obtained optimal policies. Finally, the results from the applied algorithms are compared to select a preferred algorithm for future studies. As expected, for the proposed models and strategies, a clear conflict between the health and economy performances is found. Furthermore, the results suggest that the guided usage of NPIs is  preferable as compared to refraining from employing such strategies at all. This study contributes to pandemic modeling and simulation by providing a novel concept that elaborates on integrating economic aspects while exploring the optimal moment to enable NPIs.

\end{abstract}

\section{Introduction}
This study is motivated by the current SARS-CoV-2 pandemic and the need to support policy-making on controlling the virus spreading. In particular, this study is based on the understanding that controlling the pandemic by Non-Pharmaceutical Interventions (NPIs), such as lockdowns, comes with an economic cost that creates an unavoidable dilemma to be resolved by the Decision-Makers (DMs) \cite{fernandes2020economic}. This is hereby referred to as the Health-Economy Dilemma (HED) \cite{philipson2000economic}. 

A common approach to support a decision on NPI policies is to use a compartment model such as the SEIR (Susceptible, Exposed, Infectious, Recovered) model \cite{wearing2005appropriate}. This model has been used to predict the number of infected individuals over time for major epidemics \cite{britton2010stochastic}. However, most epidemiological models do not consider the economical consequences of applying NPIs. 

Epidemiological Economics (EE) focus on studying both the economic and epidemiological aspects of managing the infectious disease (e.g., \cite{philipson2000economic, eichenbaum2020macroeconomics,perrings2014merging, alvarez2020simple}). The study in  \cite{yousefpour2020optimal} was probably the first to address the HED problem. It assumes that various rate parameters of the applied pandemic model can be controlled. Viewing the problem as a bi-objective one, optimal rate-parameters were sought to influence both the economy and the health performances. This study follows some of the ideas of \cite{yousefpour2020optimal}. Yet, the focus here is on different variables. Namely, rather than searching for optimal rate values, here the search concerns the optimal time to onset the proposed control actions. Moreover, in contrast to \cite{yousefpour2020optimal}, which uses just one algorithm, here several algorithms are applied and their suitability for the problem is studied.

\par
Inspired by \cite{tang2020estimation}, we propose an extension of the SEIR model, which integrates the health and economy aspects.  
Augmented with the effects of NPIs strategies, the proposed  model  allows identifying optimal control policies for maximizing the economic performance while minimizing infections. In other words, the extended SEIR model allows studying the involved health-economy performance trade-offs, which supports resolving the HED by the DMs. 

In addition to proposing the extended SEIR model, this paper provides a study on optimal control policies (optimal NPIs). Particularly, a bi-objective optimization problem is defined to minimize extreme peaks of infection numbers and the economic damage simultaneously. This problem is solved using existing Multi-Objective Evolutionary Algorithms (MOEAs) and the extended SEIR model. The employed  algorithms, which are taken from \cite{platemo}, search for the Pareto-optimal set of control policies and the associated Pareto-front which reveals their  performance trade-offs. Finally, a comparison study is conducted to unveil the accuracy of the results as obtained by the considered algorithms. This aims to identify the preferred algorithm for future studies.   

The rest of this paper is organized as follows. The proposed extension to the SEIR model is described in Section \ref{sec2}. Section \ref{sec3} describes the considered bi-objective optimization problem.  In Section \ref{sec4},  the experimental setup is presented. Section \ref{sec5} presents the results of the simulation and optimization experiments, which are discussed in Section \ref{sec6}. Finally, the conclusions of this study are provided in Section \ref{sec7}. 

\section{Pandemic and Economic Modelling} \label{sec2}
This section presents the mathematical formulations of the proposed model. In Subsection \ref{sec:SEIR}, the extended SEIR model is introduced. Next, in Subsection \ref{sec:economic-model}, the considered macro-economic model is integrated to the extended SEIR model. 
Finally, in Subsection \ref{sec:control-policies}, the applied control policies are described.


\subsection{The Extended SEIR Model} \label{sec:SEIR}
\begin{figure}[h]
    \centering
    \includegraphics[width=\columnwidth]{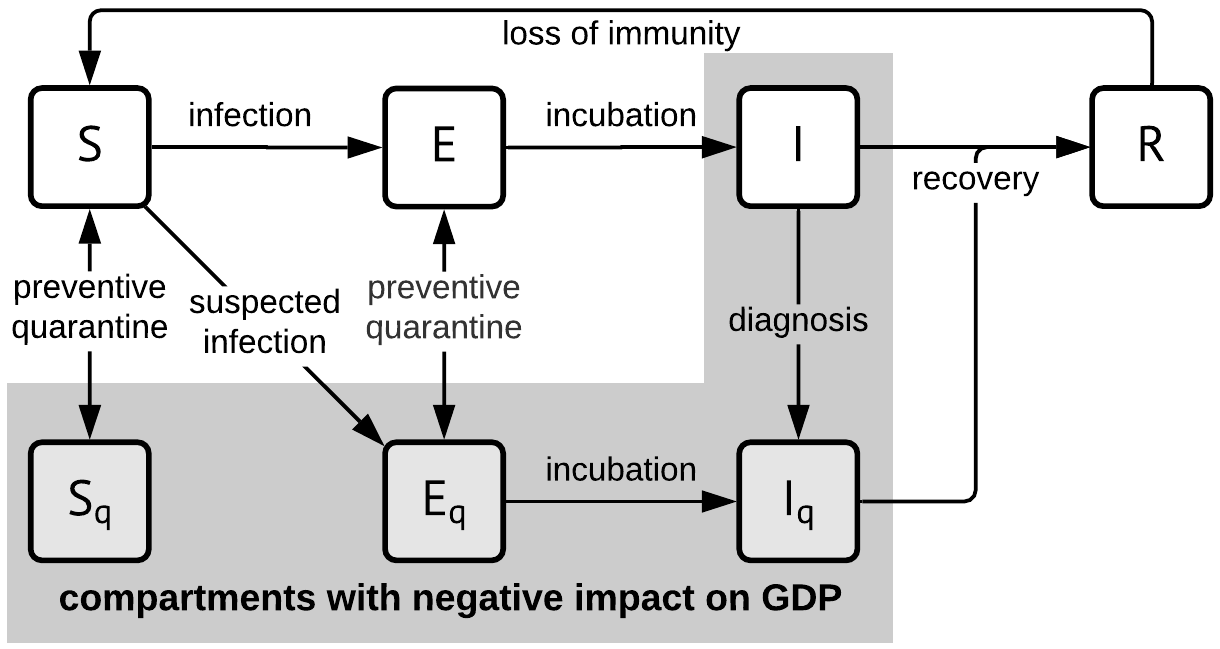}
    \caption{The extension of the SEIR model}
    \label{fig:PandemicModel}
\end{figure}

We propose a new pandemic model to support dealing with the economic consequences of having a portion of the population  quarantined during the pandemic. This portion of the population is dependent on the epidemiological status of individuals, their clinical progression, and prospective intervention measures. Specifically, we extend the SEIR compartment model in a manner inspired by \cite{tang2020estimation}. In contrast to the original SEIR model, the proposed pandemic model contains seven compartments rather than four. The considered population is divided into sub-populations (compartments), including: susceptible ($S$), susceptible quarantined ($S_q$), exposed ($E$), exposed quarantined ($E_q$), infectious ($I$), infectious quarantined ($I_q$) and recovered ($R$). Figure \ref{fig:PandemicModel} depicts the seven compartments and their relations. The four compartments, which are highlighted by a gray background, take a part in the economic extension of the model. In other words, the union of these sub-populations could be viewed as a compartment of individuals which influence the deterioration of the economy.  

Individuals in ($S$) are those who can be infected by the virus. Individuals from ($S$) can move either to ($S_q$), ($E$), or ($E_q$). In ($S_q$) there are susceptible individuals who went into quarantine as a preventive measure. Individuals are moved to ($E$) if they have contracted the virus, but do not suspect the infection. On the other hand, ($E_q$) includes exposed individuals that are quarantined. An individual enters this compartment when exposed and being suspected as infected, or when transferred into a preventive quarantine for other reasons. Individuals of the exposed states inevitably enter the infected states. For both, ($I$) and ($I_q$) it can be said that the individuals are infectious, but only those in ($I$) can actually spread the disease, whereas those in ($I_q$) cannot. When showing symptoms while in ($I$), an individual might get diagnosed with the virus and thus will enter ($I_q$). If an individual was in ($E_q$) before, then it is assumed that infection was already suspected. Hence, such individuals go straight to ($I_q$) once the incubation period is over. Individuals from ($I$) and ($I_q$) will eventually progress to ($R$) once recovered and gained immunity from the virus. However, we consider immunity to be non-permanent and hence individuals will go back to being susceptible, ($S$), after a period of being in ($R$).

The dynamics of the extended SEIR model are represented by a set of first-order differential equations, i.e., state equations, as common to such compartment models. The equations are listed in the following. 

\begin{equation}
    \begin{aligned}
    S' ={}  & - c_r (1 - t_p) c_{dp} I S
              - c_r t_p (1 - c_{dp}) I S \\
            & - c_r t_p c_{dp} I S
              - p_{qr} S 
              + p_{qer} S_q 
              + i_{lr} R
    \end{aligned}
    \label{eq:3}
\end{equation}
\begin{equation}
  Sq' = c_r(1 - t_p)c_{dp}I S 
      + p_{qr} S
      - p_{qer}S_q  
\end{equation}
\begin{equation}
  E' = c_r t_p (1 - c_{dp})I S
     + p_{qer} E_q
     - p_{qr} E
     - i_r E
\end{equation}
\begin{equation}
 Eq' = + c_r t_p c_{dp} IS
      + p_{qr} E
      - p_{qer} E_q
      - i_r E_q 
\end{equation}
\begin{equation}
    I' = + i_r E
     - d_r I
     - i_{rr} I
\end{equation}
\begin{equation}
    Iq' = + i_r E_q
      + d_r I
      - i_{qrr} I_q
\end{equation}
\begin{equation}
   R' = + i_{rr} I
     + i_{qrr} I_q
     - i_{lr} R
\end{equation}

Ten parameters are used in the proposed pandemic model. Here, ($c_r$) refers to \emph{contact-rate}. It describes how often and how closely individuals interact with each other. The contact-rate governs how fast the virus spreads, together with ($t_p$), which is the \emph{transmission probability} when an infected person interacts with a susceptible one. Next is the \emph{contact detection probability} ($c_{dp}$). It is the likelihood by which an individual will know that she was in contact with an infected person. An individual that is aware of the contact will either go into quarantine, meaning ($S_q$) if transmission did not actually occur, or into  ($E_q$) if it did. The individual moves to ($E$) if contact was made without detection, but with transmission. With a rate of $p_{qr}$ (\emph{preventive quarantine rate}), individuals who believe to be susceptible will go into quarantine without any concrete suspicion of infection. Namely, from ($S$) to ($S_q$), or from ($E$) to ($E_q$). On the other hand, they will also leave quarantine again, with a rate of $p_{qer}$ (\emph{preventive quarantine end rate}). Immune individuals in ($R$) might return to being susceptible ($S$), according to $i_{lr}$, the \emph{immunity loss rate}. Individuals will move from ($E$) and ($E_q$) into ($I$) and ($I_q$) according to the \emph{incubation rate} ($i_r$). Based on $d_r$, the \emph{diagnosis rate}, infected individuals are identified and moved from ($I$) to ($I_q$). Finally, infected individuals recover either with the \emph{infected recovery rate} ($i_{rr}$), or with the \emph{infected quarantined recovery rate} ($i_{qrr}$).

In the equations, the term $c_r (1 - t_p) c_{dp} I S$  corresponds to individuals that had contact with the infected ones but did not get the infection. They know they were in contact and thus move to ($S_q$). The term  $c_r t_p (1 - c_{dp}) I S$ corresponds to individuals who contracted the virus without suspicion and move to ($E$). The term $c_r t_p c_{dp} I S$ concerns the infected individuals with a suspicion of contact who move to ($E_q$). The term $p_{qr} S$ involves those that are susceptible and are sent to preventive quarantine ($S_q$), without any concrete suspicion. The term $p_{qr} E$ includes individuals who move into preventive quarantine without concrete suspicion, but they are actually exposed and hence move to ($E_q$). The term $p_{qer} S_q$ describes susceptible individuals who leave the preventive quarantine, hence returning to ($S$). The term $p_{qer} E_q$ represents the exposed individuals which are assumed to be unaware of their infection. These individuals are shifted back to ($E$). The terms $i_r E$ and $i_r E_q$ describe the movement of exposed individuals to the corresponding infected compartments, ($I$) and ($I_q$), following the incubation period of the virus. The term $d_r I$ represents infectious individuals who are tested, correctly diagnosed, and finally quarantined to ($I_q$). The term $i_{rr} I$ and $i_{qrr} I_q$ involve individuals which recovered from the disease and gained immunity, thus moving to ($R$). Finally, $i_{lr} R$ are individuals who lost their immunity and go back to being susceptible ($S$). The total population $N$ is assumed to be static and equals the sum of all the compartments: 

\begin{equation}
   N = S + S_q + E + E_q + I + I_q + R
   \label{eq:N}
\end{equation}

\subsection{The Economic Model} \label{sec:economic-model}
In a pandemic scenario, the risk of economic depression is high. Applying actions such as lockdowns have a negative impact on the economic activities. Generally, such pandemic control actions will deteriorate the economy and will lead to a loss of wealth \cite{sands2016assessment}. Thus, the proposed model accounts for the pandemic influence on the economy.  Specifically, we added an equation to represent the influence of the model compartments on the $GDP$. It is to be viewed as a relative value with a neutral zero baseline, starting at the beginning of the pathogen's diffusion. It will then develop positively based on an assumed innate growth of the $GDP$, and negatively based on the status of the pandemic compartments. The relation between economy and pandemic is assumed to be unilateral in this model. This means that the economic model is influenced by the pandemic's course, but not vice versa.

The current economic model assumes that no forced quarantines are implemented. This means that all individuals in ($S$), ($E$), and ($R$) act without any significant restrictions and therefore will maintain their usual economic behavior. This appears to be possible because all these individuals are symptom-free, not quarantined, and believe themselves to be uninfected, even if that is not actually true for individuals in ($E$). On the other hand, ($S_q$), ($E_q$), ($I$), and ($I_q$) have a considerable impact on the economy. This is because the concerned individuals are either quarantined, which translates to inhibited spending and productivity, or infected, thus their contribution to the economy is reduced noticeably. Next, the following new compartment ($C_e$) is introduced to reflect individuals that influence the economy. 
\begin{equation}
    C_e=p_{qi}S_q+e_{qi}E_q+i_iI+i_{qi}I_q \label{eq:12}
\end{equation}
The economy compartment influence the $GDP$ according to the  following differential equation:
\begin{equation}
    GDP'=b_g-p_i(C_e) \label{eq:11}
\end{equation}

The above equation involves six parameters, which brings the total amount of parameters for the combined model to 16. The \emph{baseline growth} is indicated by ($b_g$) and represents the innate growth of the GDP, which we assume to be positive and linear. The other parameters concern the influence of the pandemic on the economy. The \emph{pandemic influence} ($p_i$) is a scaling factor for the overall impact of the pandemic on the economy. This allows scaling the total impact of the pandemic on the economy. The last four parameters describe how strong the negative impact of ($S_q$), ($E_q$), ($I$), and ($I_q$) are on the $GDP$. We assume that economic behavior is changing depending on the compartment, e.g., individuals in ($S_q$) might be more active as individuals in ($I_q$). Hence there is one parameter dedicated for each of these compartments: \emph{susceptible quarantined impact}, ($s_{qi}$), \emph{exposed quarantined impact}, ($e_{qi}$), \emph{infected impact}, ($i_i$), and \emph{infected quarantined impact}, ($i_{qi}$).

\subsection{Control Policies} \label{sec:control-policies}
 In the context of our model, a control policy refers to an action, intervention, or event which may influence the pandemic and the economy, and will alter their predicted trend. In the current study, two major types of actions are considered including: social distancing and lockdown.

It is suggested here to implement the policies by simple parametric adaptations. Each of the selected policies influences a single parameter within the model over a certain time frame. The influenced model-parameters are ($c_r$) for social distancing, and ($p_{qr}$) for lockdown. It should be noted that for simplicity, in our model, individuals that are locked-down are considered as in quarantine. Each of these model-parameters is effected by a time dependent influence curve. This curve is defined by the following five variables. The \emph{amplitude} defines how strong the policy will affect the parameter. At a zero amplitude, a policy would have no effect at all. Any other value will simply be added to the corresponding model parameter, individually and independently for each time step. The \emph{trigger time} defines the time stamp $t$ when the policy is going to become active. From that point in time, the policy influence will steadily rise before reaching its peak. Once reaching its peak, the influence will fade away. The extent by which each of these phases is lasting is governed by additional three variables including \emph{buildup}, \emph{peak}, and \emph{fade}. The total duration of the policy is equal to the sum of these three parameters. Describing the policy by such a time dependent influence curve allows modeling the expected delay in the public response to new regulations  and the public fatigue over time to follow the rules. In the current study all the curve-variables are kept constant except for the triggering time. 

\begin{figure}[h]
    \centering
    \includegraphics[width=\columnwidth]{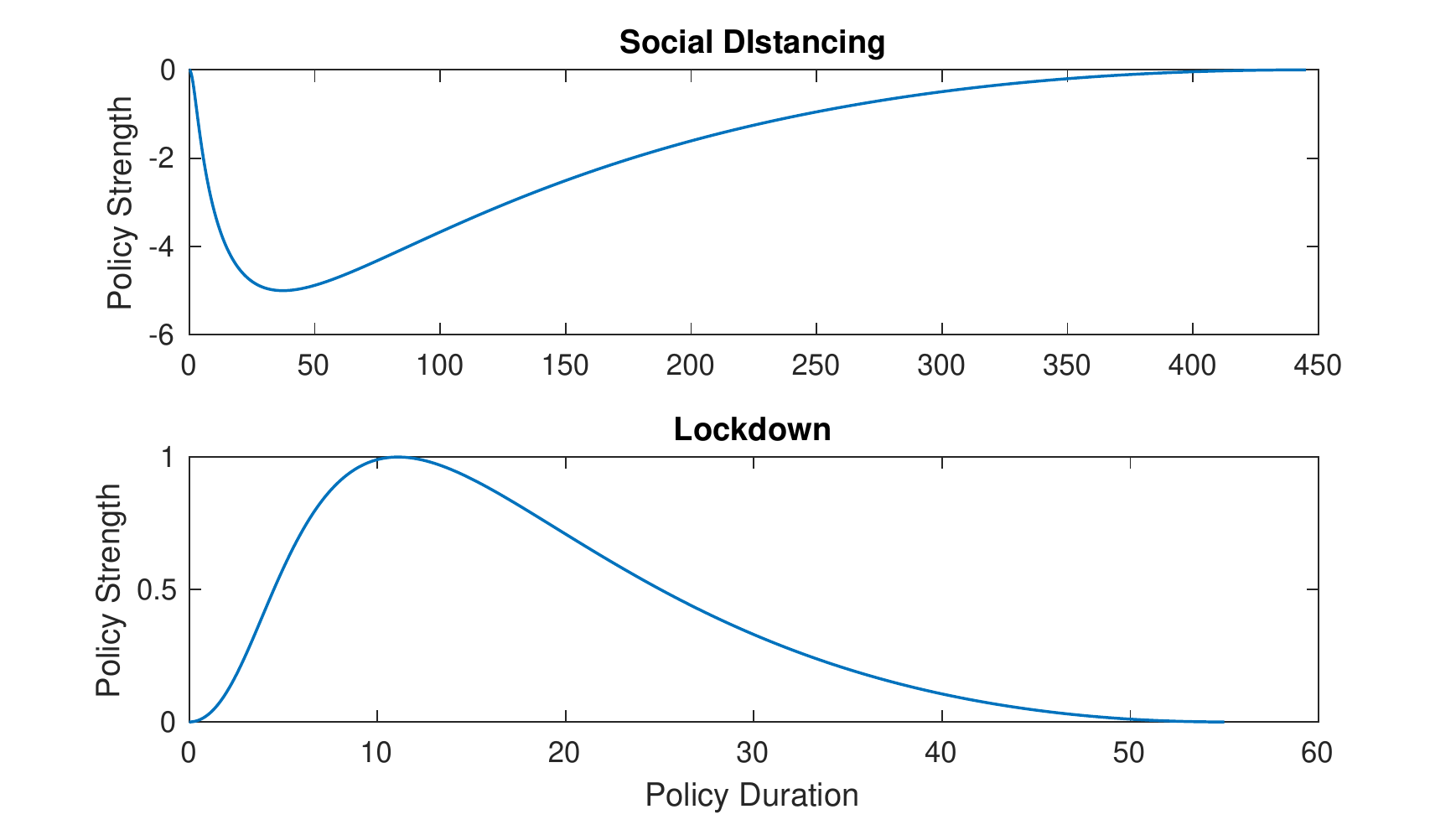}
    \caption{The influence curves for the social distancing and lockdown policies.}
    \label{fig:SocialDistancingAndLockdown}
\end{figure}

In the following simulation, Bézier curves implement the influence of the policy model. Each such curve is multiplied by a specific scaling factor, ensuring that the peak of the Bézier curve actually corresponds to the given amplitude of the policy. As such, the \textit{x}-coordinate of the curve corresponds to the time relative to the starting point of the policy, whereas the \textit{y}-coordinate corresponds to the policy's efficacy at that relative time. The control points of the Bézier curve are given as: 
\begin{equation}
    \begin{split}
        Control Point 1: (0, 0) \\
        Control Point 2: (buildup, 0) \\
        Control Point 3: (buildup, 1) \\
        Control Point 4: (buildup + peak, 1) \\
        Control Point 5: (buildup + peak, 0) \\
        Control Point 6: (buildup + peak + fade, 0) 
    \end{split}
\end{equation}

The control points define the shape of the curve.  When multiplying the  Bézier curve by the scaling factor, the policy will reach its stated amplitude in exactly one point. The scaling factor also accounts for negative amplitudes. With this scaled curve and the trigger time the policies influence is fully defined over time. Figure \ref{fig:SocialDistancingAndLockdown} shows plots of the associated scaled  Bézier curves for the policies social distancing and lockdown with the parameters as listed in Table \ref{Table:ExperimentParameters}.

\section{The Optimization Problem} \label{sec3}

Next, we formulate a bi-objective optimization problem that minimizes health and economic objectives:
\begin{equation}
    \begin{aligned}
        \mmin f_1(\textbf{t}), \mmin f_2(\textbf{t}),
    \end{aligned}
\end{equation}
\begin{equation}
    \begin{aligned}
        \textbf{\textit{where}} f_1(\textbf{t}) &= \mmax\limits_{\textbf{t}} (E(t, \textbf{t}) + E_q(t, \textbf{t}) + I(t, \textbf{t}) + I_q(t, \textbf{t})) \\
        \textbf{\textit{and}}  f_2(\textbf{t}) &= - \mmin\limits_{\textbf{t}} GDP(t, \textbf{t})
    \end{aligned}
\end{equation}
where $f_1$ corresponds to the health objective and $f_2$ corresponds to the economic objective. The goal of $f_1$ is to minimize the peak of concurrent amount of infected or exposed individuals  (\textit{flatten the curve}). The objective of $f_2$ is to avoid or minimize a recession of the economy. If $f_2$ is optimal ($f_2 = 0$), the $GDP$ will never be lower than at the beginning of the observed period. Here, $t$ refers to the specified time stamp, where $t \in \mathbb{R} : 0 \leq t \leq t_{max}$ and $t_{max}$ indicates the last observed time stamp. In our experiments we define $t_{max} = 300$. The decision vector is denoted as $\textbf{t}$. It contains the triggering times for social distancing and lockdown, which are the optimization variables of the current study. 

According to the Pareto-optimality approach, both of these objectives are considered simultaneously without any a-priori articulation of the objective preferences. Given the conflicting objectives, this results in a set of Pareto-optimal solutions and their associated front in the objective-space. The proposed solution approach provides trade-off information, which is expected to support a rational strategy selection.

\section{Experimental Setup} \label{sec4}

In this section, the experimental setup is presented. All simulations are performed using MATLAB R2019b on a MacBook Pro with 2.2 GHz Intel Core i7 processor and 16 GB of 2400 MHz DDR4 RAM. The employed parameters are shown in Table \ref{Table:ExperimentParameters}. For the optimization study, four algorithms have been used including: NSGA II (Non-dominated Sorting Genetic Algorithm II) \cite{NSGAII}, NSGA III (Non-dominated Sorting Genetic Algorithm III) \cite{NSGAIII}, MOPSO (Multi-Objective Particle Swarm Optimization) \cite{MOPSO}, and MOEA/D (Multi-Objective Evolutionary Algorithm based on Decomposition) \cite{MOEAD}. Such MOEAs are designed to solve multi-objective problems by searching for the Pareto-optimal solutions and their associated front. Unless otherwise stated, these algorithms are implemented using their codes in PlatEMO \cite{platemo} with their default settings. 

For each algorithm and run, we used a population size of 100 and 4,000 evaluations to find optimal solutions. For this parameter setting, the trigger time's upper bound is set to 100 since we assume that there will be no appropriate solutions when on-setting the policies later. Due to the stochastic nature of the employed algorithms, 36 independent runs were performed for each algorithm to reach statistical conclusions. To compare the algorithm performances, we  produced a reference front and an associated reference solution set by combining all of the individual Pareto fronts from all runs and by the removal of all dominated solutions (see Section \ref{sec5}).   

\begin{table}[h]
	\centering
	\caption[Experiment Parameters.]{Parameters used in the experiments.} \label{Table:ExperimentParameters}
	\begin{tabular}{ll} 
		\hline
		Parameter & Value\\
		\hline
		\multicolumn{2}{c}{\textit{Extended SEIR Model Parameters}}\\
		Initial $S$                         & 0.98\\
		Initial $E$, $I$                    & 0.01\\
		Initial $S_q$, $E_q$, $I_q$, $R$    & 0\\
		Contact Rate $c_r$                  & 10\\
		Transmission Probability $t_p$      & 0.1\\
		Incubation Rate $i_r$               & 1/7\\
		Preventive Quarantine Rate $p_{qr}$ & 0\\
		Contact Detection Probability $c_{dp}$ & 0.05\\
		Diagnosis Rate $d_r$                      & 1/14\\
		Infected Recover Rate $i_{rr}$               & 1/14\\
		Infected Quarantined Recover Rate $i_{qrr}$   & 1/14\\
		Immunity Loss Rate $i_{lr}$                 & 1/90\\
		\hline
		\multicolumn{2}{c}{\textit{Economy Model Parameters}}\\
		Initial $GDP$                   & 0\\
		Baseline Growth $b_g$           & 0.02\\
		Pandemic Influence $p_i$             & 0.12\\
		Preventive Quarantined Impact $p_{qi}$   & 0.4\\
		Exposed Quarantined Impact $e_{qi}$      & 0.4\\
		Infected Impact $i_i$                 & 0.8\\
		\hline
		\multicolumn{2}{c}{\textit{Fixed Policy Parameters}}\\
		Social Distancing Amplitude     & -5\\
		Social Distancing Buildup       & 5\\
		Social Distancing Peak          & 40\\
		Social Distancing Fade          & 400\\
		Lockdown Amplitude              & 1\\
		Lockdown Buildup                & 5\\
		Lockdown Peak                   & 10\\
		Lockdown Fade                   & 40\\
		\hline
		\multicolumn{2}{c}{\textit{Optimization setup.}}\\
		Population Size                     & 100\\
		Evaluations                         & 4,000\\
		Runs                                & 36\\
		Decision Variables Upper Bound      & 100\\
		Decision Variables Lower Bound      & 0\\
		\hline
	\end{tabular}
\end{table}

\section{Results} \label{sec5}
This section presents the results of the simulation and optimization experiments: Based on the defined objectives, we optimize the  control policy's  trigger times to find optimal trade-off solutions between the health and economic objectives. We compare NSGA II, NSGA III, MOEA/D, and MOPSO  to find possible strategies for activating the control policies (social distancing and lockdown). 
\begin{figure}[h]
    \centering
    \includegraphics[width=\columnwidth]{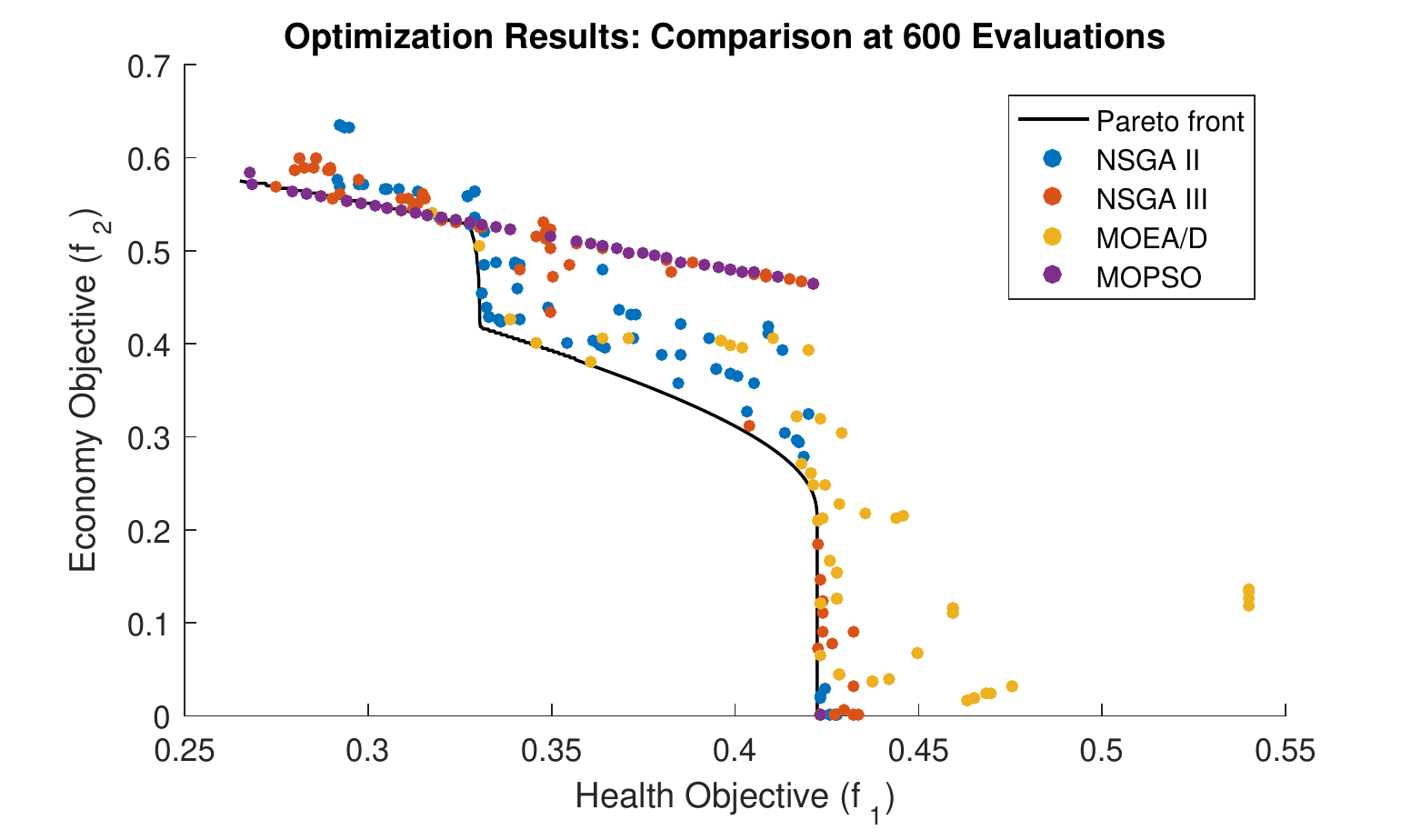}
    \caption{Comparison of four multi-objective optimization algorithms after 600 evaluations each.}
    \label{fig:Comparison600}
\end{figure}

\begin{figure}[h]
    \centering
    \includegraphics[width=\columnwidth]{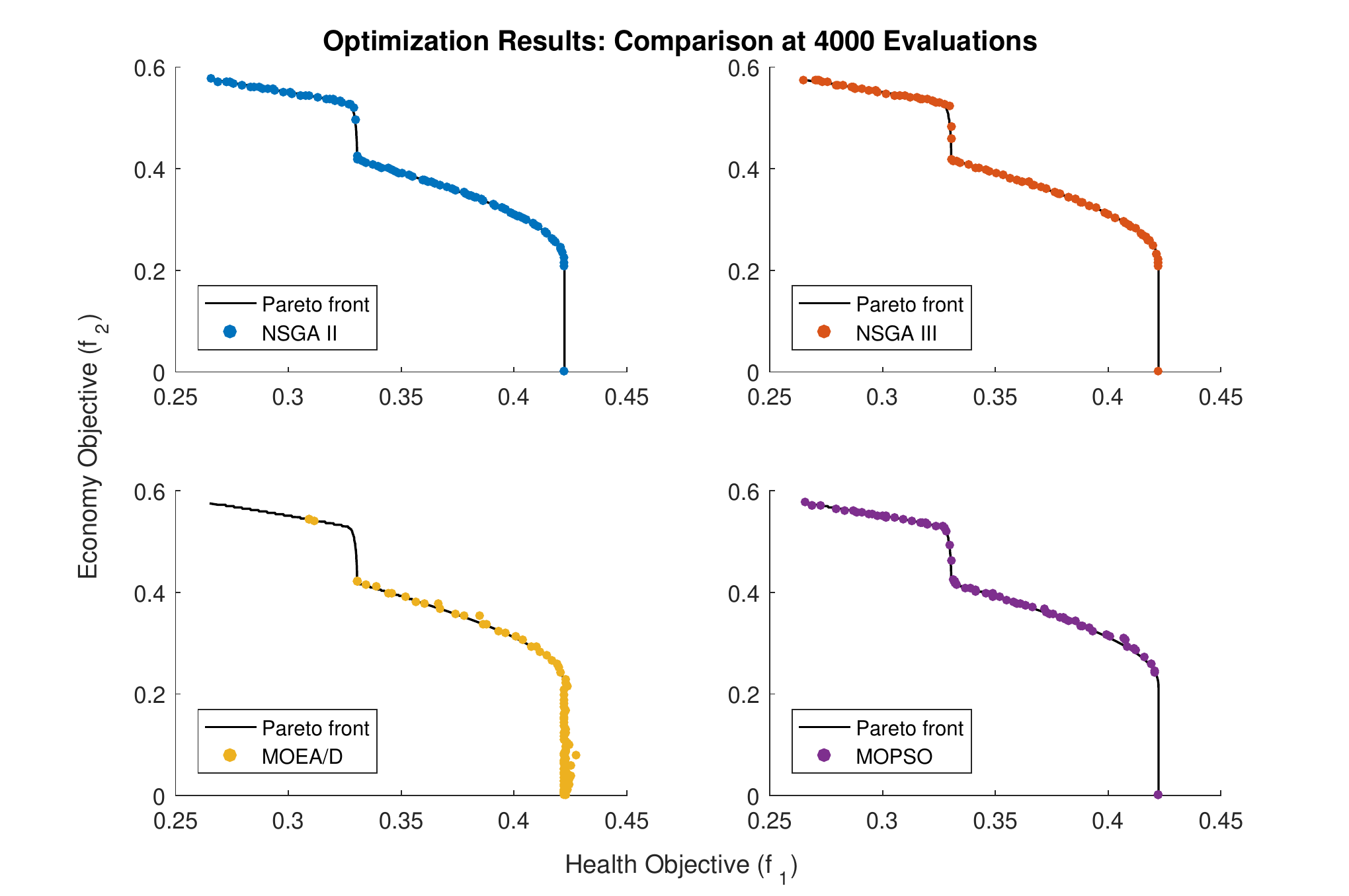}
    \caption{Comparison of four multi-objective optimization algorithms after 4,000 evaluations. A single, randomly chosen run is plotted for each algorithm.}
    \label{fig:MatrixPlotComparison4,000}
\end{figure}


\begin{figure}[h]
    \centering
    \includegraphics[width=\columnwidth]{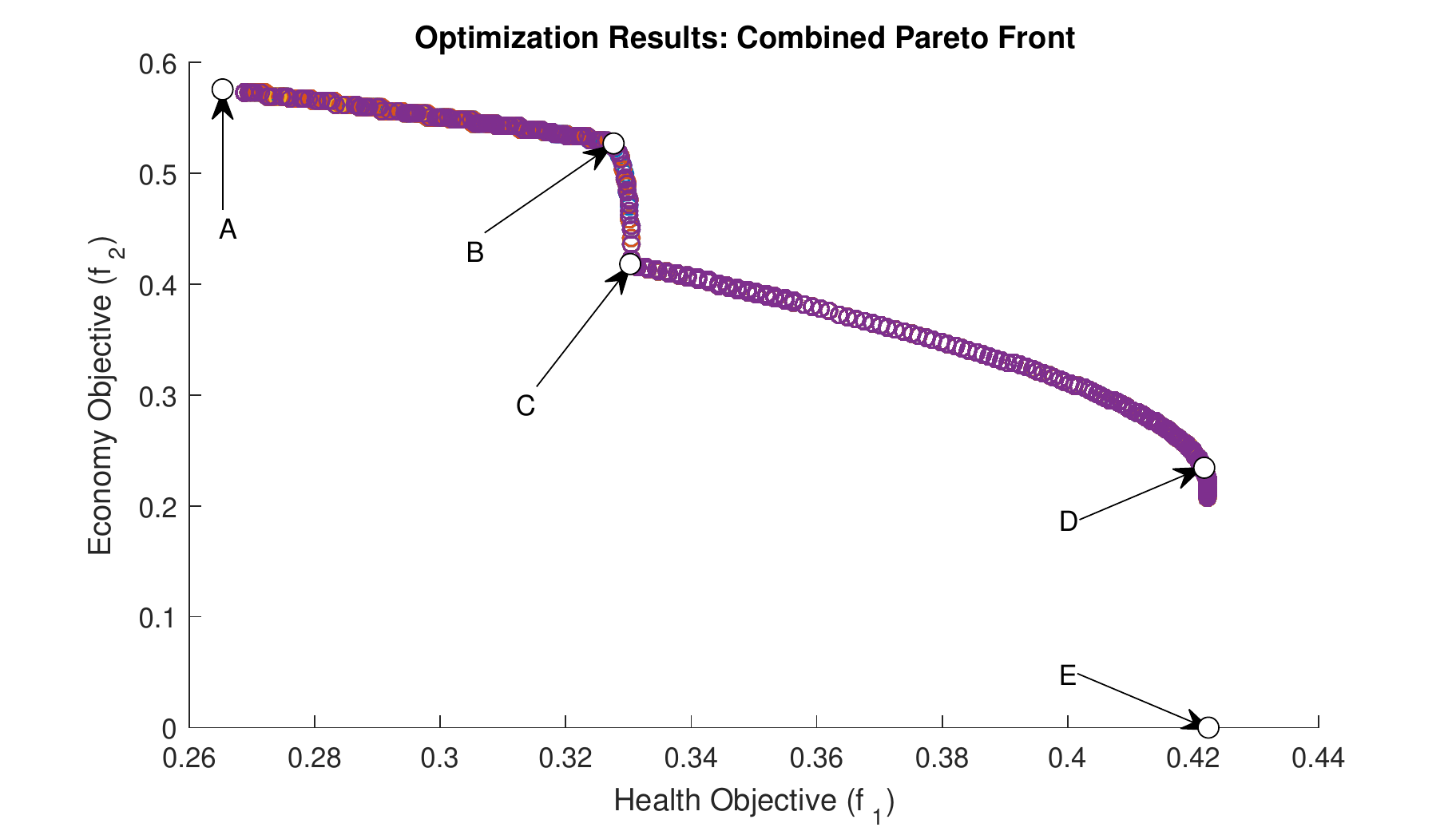}
    \caption{Combined Pareto front of four multi-objective optimization algorithms over 36 runs with 4,000 evaluations each.}
    \label{fig:ParetoFront}
\end{figure}

\begin{figure}[h]
    \centering
    \includegraphics[width=\columnwidth]{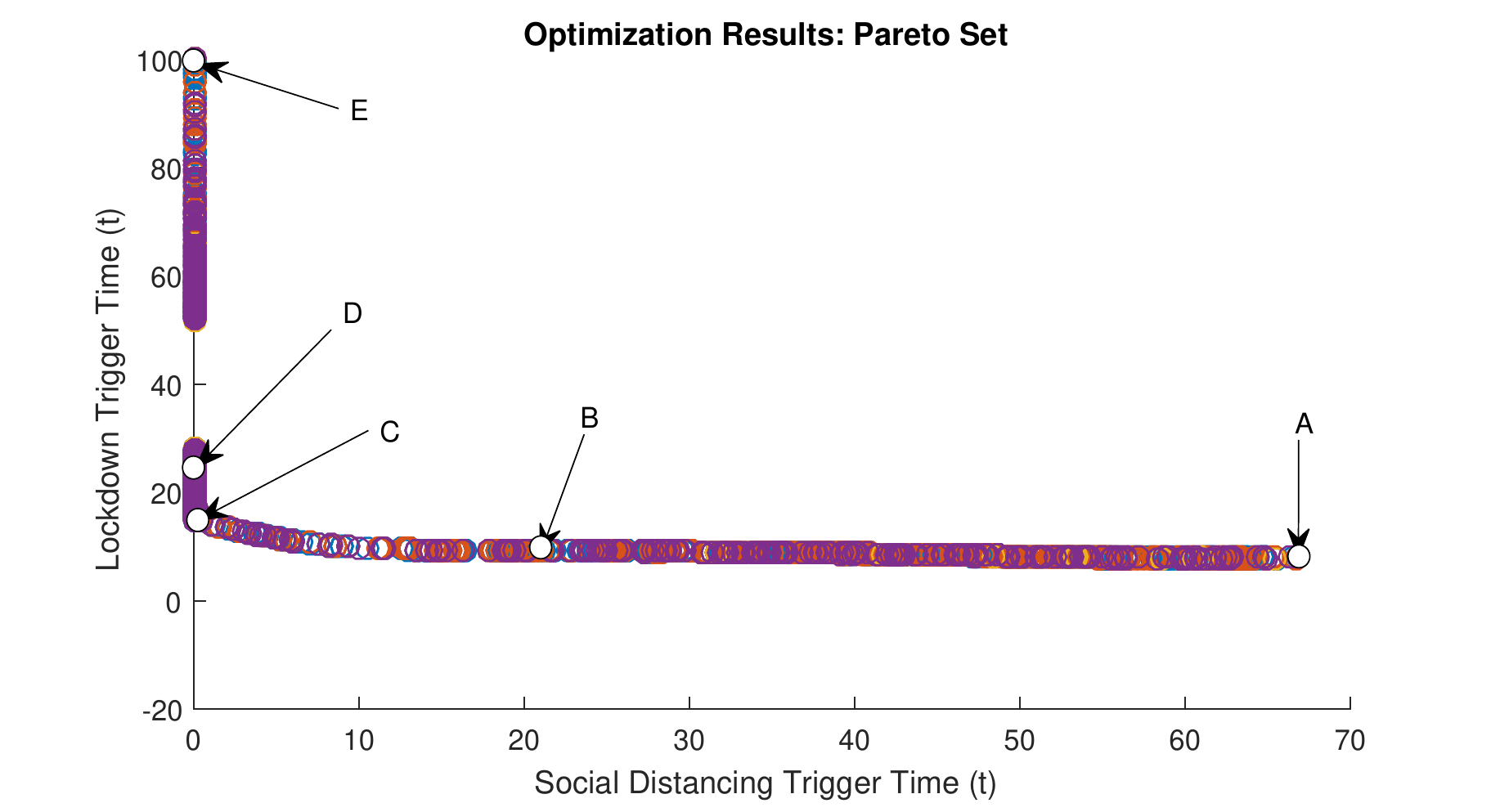}
    \caption{Combined Pareto set corresponding to the solutions displayed by the combined Pareto front.}
    \label{fig:ParetoSet}
\end{figure}

Figure \ref{fig:Comparison600} illustrates some  vectors as obtained by the considered algorithms after 600 evaluations. Also shown is a curve, which can be viewed as an estimation of the Pareto-front. In view of this curve, it can be observed that none of the  algorithms was able to reach a good representation of the entire front within 600 evaluations. Next, Figure \ref{fig:MatrixPlotComparison4,000} shows, for each algorithm, the resulting front after 4,000  evaluations. Observing the figure, it can be concluded  that all the algorithms reached at least some part of the reference front after 4,000 evaluations. These results should be evaluated when compared with the reference front, which is shown in Figure \ref{fig:ParetoFront}. The conclusions from both figures is that MOEA/D failed to find a good representation of the front, whereas the other three algorithms did. It is particularly noticeable that it lacks diversity, with many solutions clustered around the lower right part of the front. It should be noted that the points, which were found by MOEA/D, along the vertical line at $f_1$=0.42 are dominated by the lowest point in the reference front. 

In Figure  \ref{fig:ParetoFront}, five points are highlighted, indicated by the labels \textit{A, B, C, D,} and \textit{E}. These were selected at the edges of each of the apparent regions of the reference front. Each of these points is associated with a particular optimal strategy, as shown in Figure \ref{fig:ParetoSet}. The shape of the combined front in Figure \ref{fig:ParetoFront} strongly suggests a significant trade-off that is inherent to the presented HED problem. Generally, no strategy can simultaneously satisfy the health and economic objectives to a large extend. Moreover, the obtained front is not convex. In fact, the shape of the front indicates that if the traditional weighted-sum solution approach was used, the obtained front would have resulted in the two extreme points of the front, hence hiding most of the potential alternatives.  


 Table \ref{Table:PerformanceIndicators} aims to further highlight the differences between the performances of the employed algorithms. It shows the statistics for three major performance indicators including: Hyper-Volume (HV), Inverted Generational Distance (IGD), and Spread (SD). The statistical significance was tested with a Mann-Whitney U test and \textit{p-value} for all preceding Kruskal-Wallis tests is $0.0$. 
\begin{table}[h]
\addtolength{\tabcolsep}{0pt}
	\centering
	\caption[Algorithm Performance Indicators.]{Mean (standard deviation) Performance Indicators at 4,000 Evaluations.} \label{Table:PerformanceIndicators}
	\resizebox{\columnwidth}{!}{%
\begin{threeparttable}
	\begin{tabular}{llll} 
		\hline
		Algorithm & HV & IGD & SD \\
		\hline
		NSGA II     & \textbf{2.0E-01 (7.8E-05)} & \textbf{1.4E-03 (1.1E-04)} & 7.6E-01 (8.9E-02)\textdagger \\
		NSGA III    & \textbf{2.0E-01 (1.7E-04)}\textdagger & \textbf{1.4E-03} \textbf{(1.4E-04)}\textdaggerdbl & \textbf{1.0E+00} (\textbf{1.5E-01})  \\
		MOEA/D      & 1.8E-01 (7.6E-03)\textdagger  & 1.9E-02 (1.2E-02)\textdagger  & 6.8E-01 (1.1E-01)\textdagger  \\
		MOPSO       & 1.9E-01 (5.8E-03)\textdagger & 5.1E-03 (1.4E-02)\textdagger & 7.5E-01 (1.0E-01)\textdagger \\
		\hline
	\end{tabular}
	\begin{tablenotes}
	\item[] The NSGA II (HV and IGD) or NSGA III (SD) performs significantly (p-value $\leq 0.01$) better (\textdagger) and equivalent (\textdaggerdbl), respectively, in comparison with the corresponding algorithm. 
        \end{tablenotes}
\end{threeparttable}
}
\end{table}

In case of HV, NSGA II performed significantly better than all other algorithms. When evaluating IGD, NSGA II and NSGA III performed significantly better than the other algorithms, while showing no significant difference between each other. Considering SD, NSGA III performed significantly better than all other algorithms. Overall, NSGA II and NSGA III performed best, showing both good convergence and spread of the solutions along the Pareto-front. MOPSO also found good solutions, whereas MOEA/D performed the worst out of the four algorithms.

\begin{figure}[h]
    \centering
    \includegraphics[width=\columnwidth]{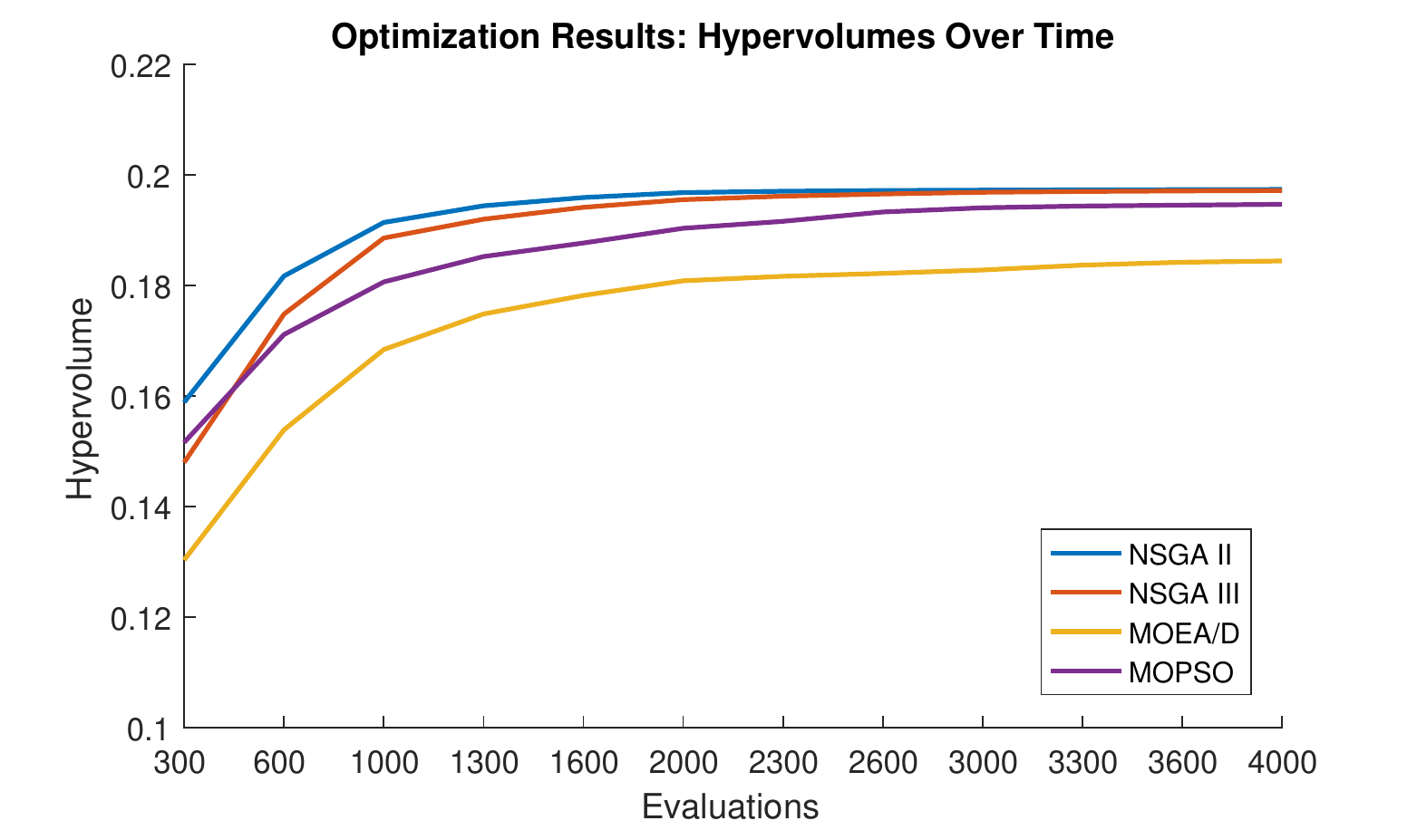}
    \caption{The HV averages vs. evaluations.}
    \label{fig:Hypervolumes}
\end{figure}

Figure \ref{fig:Hypervolumes} provides a comparison of the convergence performance of the four optimization algorithms. It shows the evolution of their HV with respect to increasing number of evaluations. The shown curves corresponds to the combined Pareto front from all the runs of each algorithm.  Evidently, NSGA II and NSGA III converged faster and reached higher values, as compared with the other two algorithms. MOPSO seems to have convergence difficulties, which can also be observed when viewing the results in Figure \ref{fig:Comparison600}. The stagnating convergence of MOEA/D suggests that its performance would not change significantly with additional evaluations.

\section{Strategy Evaluation } \label{sec6}

The marked strategies in Figure \ref{fig:ParetoSet}  are discussed in the following. The discussion assumes that the ratio objective-space's scales reflects trade-offs of interest to the DMs (i.e., shifting by 0.02 in the health performance is comparable in some way to shifting by 0.1 in the economy performance). Comparing strategies A and B, it can be observed that in both strategies, lockdown is triggered in an early stage. These strategies differ primarily by the triggering time of the other action, i.e., social distancing.  The associated performance of A is the best in terms of health but the worst in terms of economy. When triggering social distancing earlier, as done in strategy B, the economy performance is improved at the cost of deteriorating the health performance. A reversed situation exists when comparing strategies E and D. Both involves triggering the social distancing at the earliest possible time. Strategy E provides the best economy performance but the worst health performance. When shifting from E to D, there is almost no gain in the health performance as compared with the loss in the economy. Finally, its worth to compare strategy C with strategies B and D. When shifting from B to C, the health cost appears small as compared with the economy gain. When shifting from D to C there is also a trade-off, which might be worthwhile to consider.  It should be noted that there is no single optimal strategy of triggering both policies late, which appears logical. The following provides a general description about the dynamics of the pandemic compartments and the GDP, as obtained by alternative strategies including a baseline strategy of not acting at all. 

\subsection{Strategy I: No policies}

\begin{figure}[h]
    \centering
    \includegraphics[width=\columnwidth]{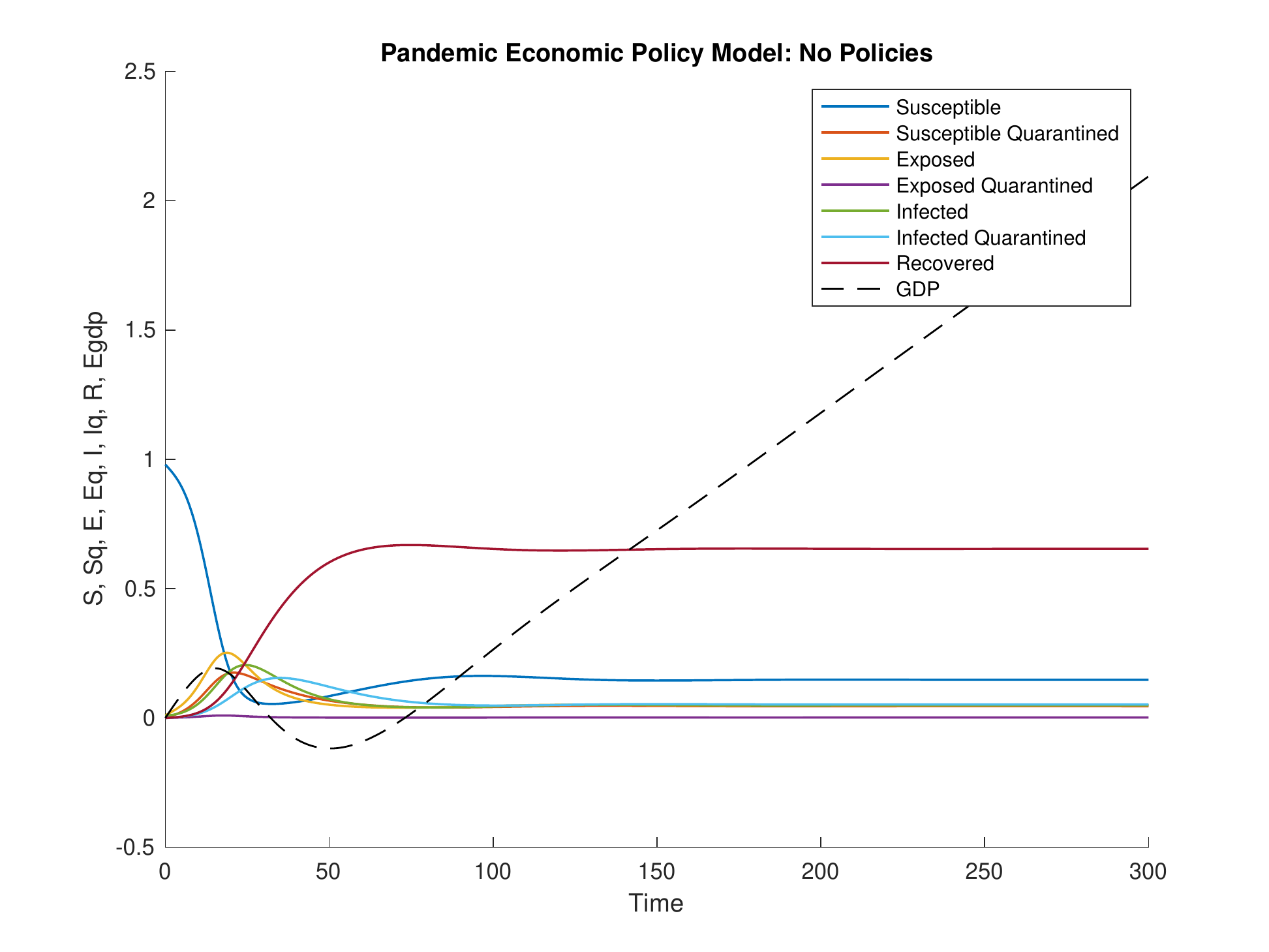}
    \caption{No control strategy at all}
    \label{fig:NoPolicies}
\end{figure}

The first strategy represents the baseline of what would have happened if no policies were applied. As Figure \ref{fig:NoPolicies} shows, this approach quickly develops herd immunity while sustaining economic health. However, the public-health system suffers in this no-strategy scenario, which causes many fatalities, and may result in overloading and possibly a collapse of the system. In the studied case, during the pandemic, a maximum of more than 54\% ($f_1 = 0.5403$) of the population would be infected and the $GDP$ would decline by $0.1178$.

\subsection{Strategy II: Health-based policy}
\begin{figure}[h]
    \centering
    \includegraphics[width=\columnwidth]{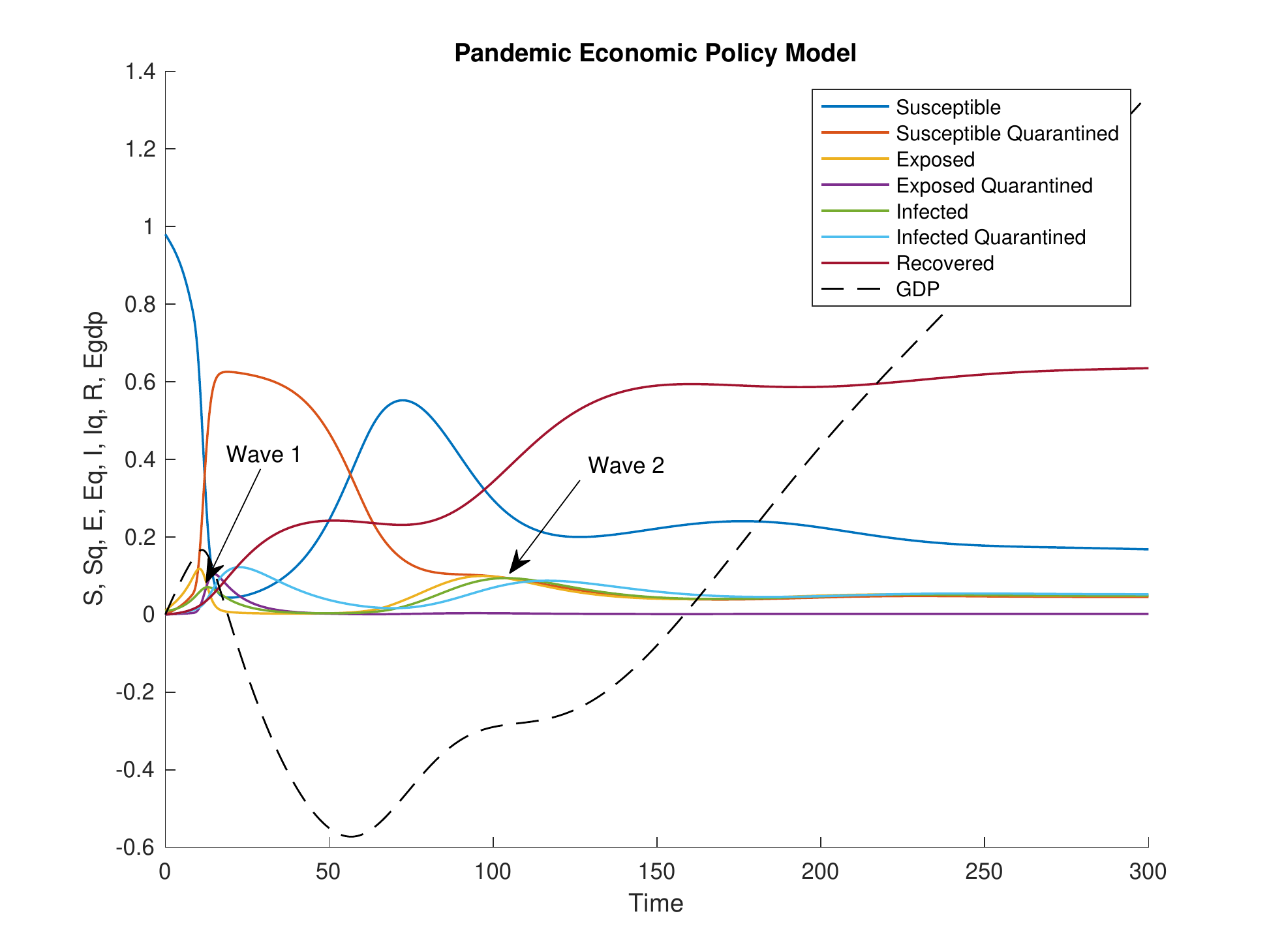}
    \caption{Optimal strategy for minimizing the health objective.}
    \label{fig:OptimalHealth}
\end{figure}

Next, we consider a case where policymakers prioritize health over economy. In such a case, the optimization suggests an early lockdown, followed by late social distancing. Figure \ref{fig:OptimalHealth} illustrates the dynamics of the process in the case of setting the triggering of the lockdown to be  at $t=8.00247$ and for the social distancing at $t=66.8182$. This corresponds to the highlighted solution A from Figures \ref{fig:ParetoFront} and \ref{fig:ParetoSet}. As can be observed form Figure \ref{fig:OptimalHealth}, these points in time correspond roughly to the beginning of the pandemic's first and second waves. It appears that the chosen strategy managed to reduce the impact of those waves. There is a maximum of 26.5\% ($f_1 = 0.2650$) infected individuals at a specific point in time. In this scenario, the economy declines significantly ($f_2 = 0.5752$) and stays depressed for relatively long time. Figure \ref{fig:OptimalHealth} also shows that herd immunity develops much later as compared with the first (baseline) strategy of no action at all, while having around 50\% less simultaneously infected individuals. 
\subsection{Strategy III: Economy-based policy}
This strategy focuses on minimizing economic losses while considering health as a secondary objective. It is characterized by a very early social distancing, triggering already at $t=0.09506$, and a very late lockdown, triggering at $t=100$. The solution presented here is also highlighted as point E in Figures \ref{fig:ParetoFront} and \ref{fig:ParetoSet}. While this strategy achieves the highest possible objective value of $f_1 = 0$ (the economy would not decline), it neglects the health objective and allows  42.23\% of individuals to be infected simultaneously. 

\begin{figure}[h]
    \centering
    \includegraphics[width=\columnwidth]{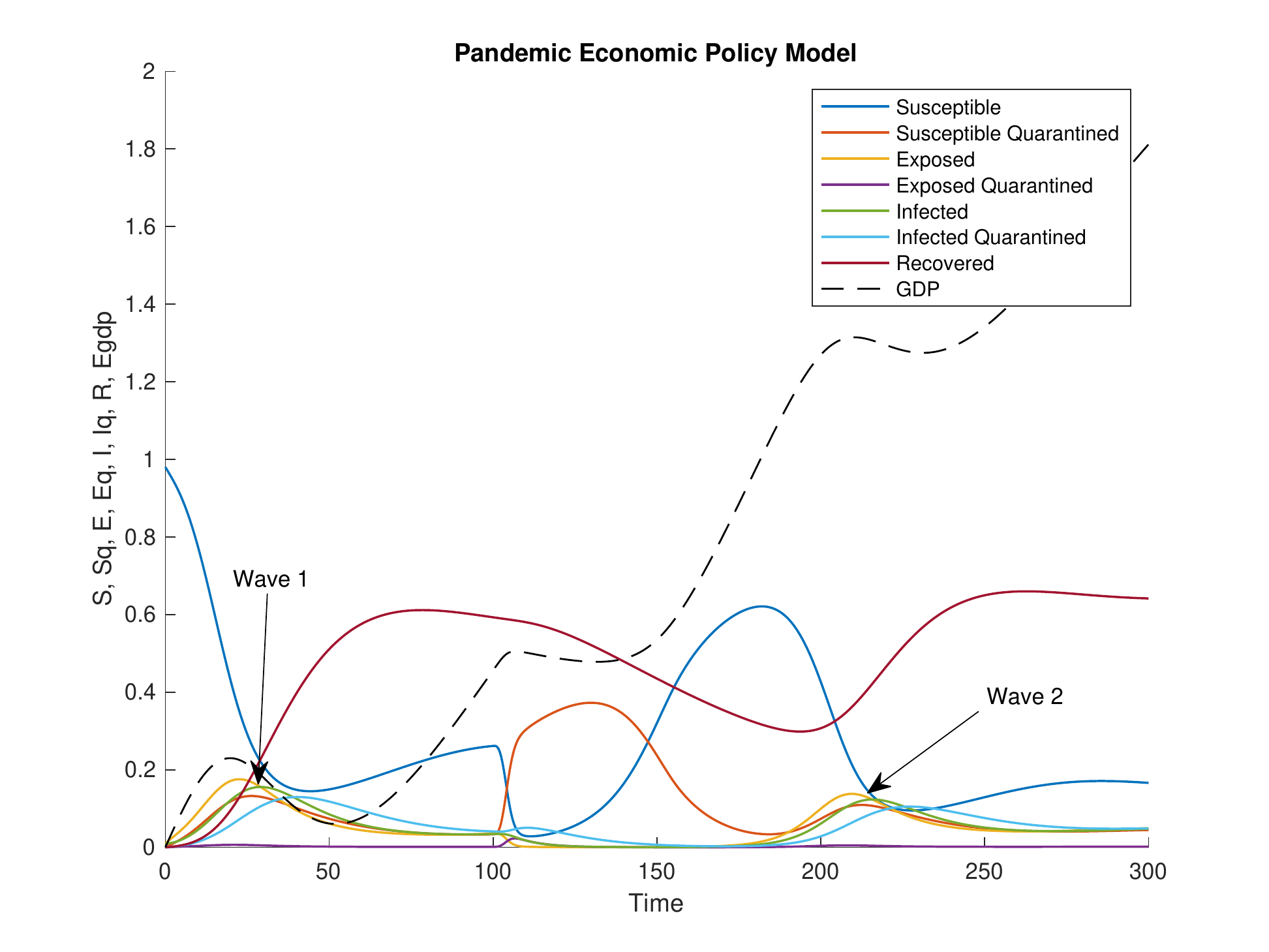}
    \caption{Optimal strategy for maximizing the economy benefits.}
    \label{fig:OptimalEconomy}
\end{figure}

Still, this strategy offers better performances in both objectives, when compared with the baseline condition (no policies). Figure \ref{fig:OptimalEconomy} shows that this strategy would delay herd immunity significantly. We also observe that our model's constraint to use each policy exactly once can be counterproductive. Without this rule, there would have been herd immunity around time step 100, but the second policy still needed to be used, triggering a second infection wave. Future studies should investigate this dynamic more closely.

\subsection{Strategy IV: A trade-off policy}

Finally, we present an optimal trade-off strategy between health and the economy. 

\begin{figure}[h]
    \centering
    \includegraphics[width=\columnwidth]{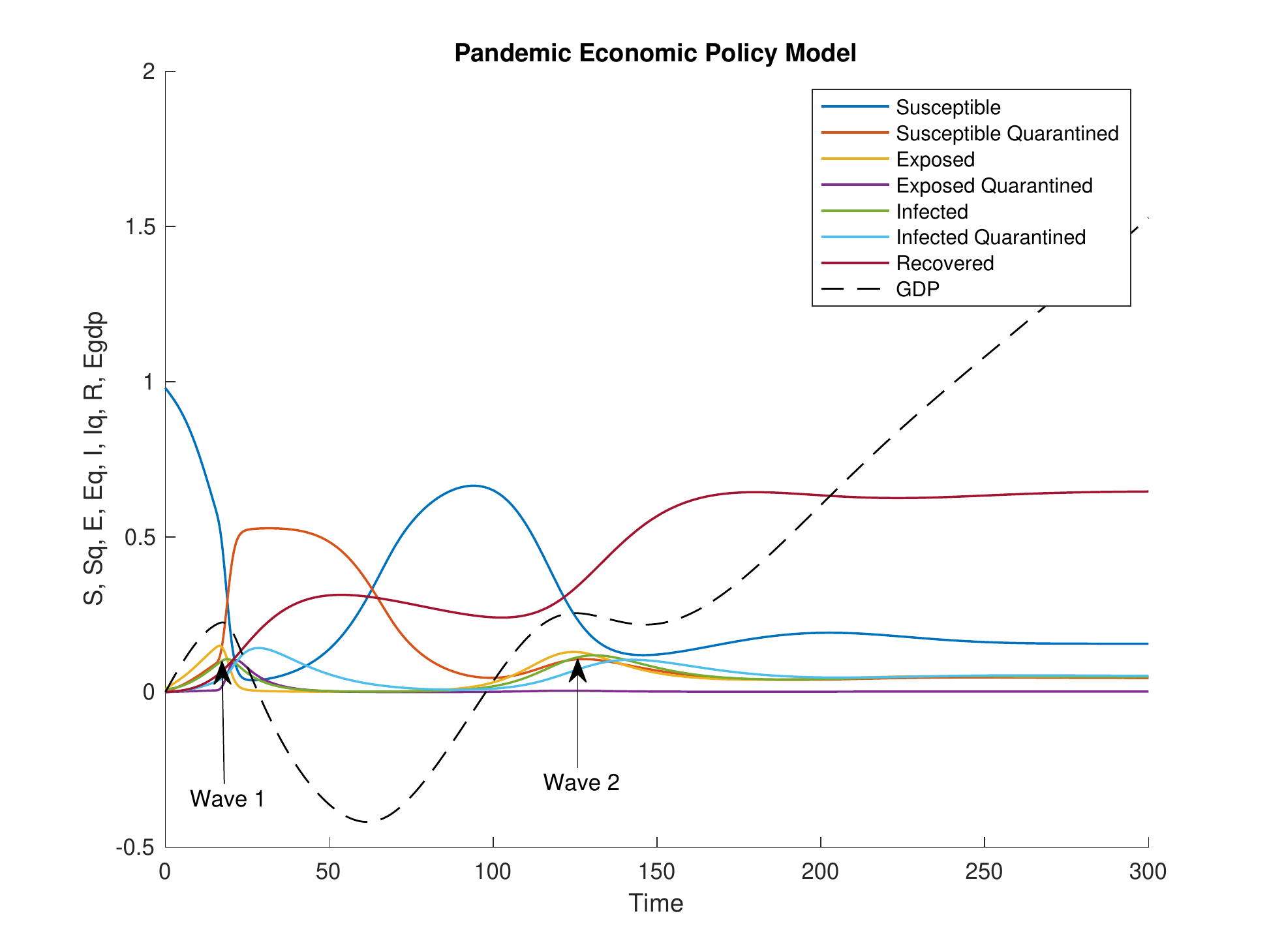}
    \caption{Optimal trade-off strategy.}
    \label{fig:OptimalTradeoff}
\end{figure}

For the chosen strategy, both actions are in place comparatively early. The social distancing first, at a time of $t=0.0584$, followed by the lockdown on time step $15.0575$. Figure \ref{fig:OptimalTradeoff} shows a reasonably good health objective of $f_1 = 33.06\%$ maximal simultaneously infected individuals, while the economy losses reached a point below the baseline at $f_2 = 0.4181$. Comparing this strategy to the  two strategies of the above subsections, it is obvious that it cannot compete with their strong focus on a single objective, but concurrently it does not neglect any objective as much as both of them did. 

\section{Summary and Conclusions} \label{sec7}
This study proposes an extended macro-level pandemic model that includes an economic perspective. The model allows examining the impact of various policies on both the economy and the pandemic spread. Next, a bi-objective optimization problem is presented with conflicting health and economy objectives. The proposed model is used to search for the Pareto-optimal strategies, for the aforementioned problem. In this study, each strategy is composed of triggering times  for social distancing and lockdown. To solve the presented problem, four multi-objective optimization algorithms are used including: NSGA II, NSGA III, MOPSO, and MOEA/D. NSGA II and NSGA III provided significantly better results than the two other algorithms. Furthermore, this study includes a discussion on various strategies and their trade-offs, as revealed by the optimization. As expected, these strategies provide better performances as compared with no action at all. What stands out in the current study is that a second wave might be inevitable once the effects of the applied policies fade. Since that no pharmaceutical interventions is considered here, only herd immunity can  end the pandemic in the proposed model. 

Future research may extend the presented model in many ways. For example, explicit compartments might be added for asymptomatic infections, including the number of case fatalities, the vaccination rate, or effects on mental health. Also, implementing sector-specific curves can extend the economic model, e.g., travel, agriculture, or services. Financial support and long-term effects on the economy are further points to take into consideration. It is noted that the current study focuses on two types of control policies, whereas many more can be suggested. Furthermore, the policies' strength and duration should be optimized as well. Apart from these, the considered objectives could be extended. For example by considering both long and short term impacts not just on the spread of the pandemic and the economy, but also on aspects such as mental health and education. 

\section*{Acknowledgement}
The authors would like to acknowledge the support of \textbf{The Volkswagen Foundation} to carry out this research. The authors would also like to thank Prof. Bruria Adini and Prof. Leonardo Leiderman from Tel Aviv University, Israel, and Prof. Ingrid Ott from Karlsruhe Institute of Technology, Germany, for fruitful discussions on the topic of this study.

\bibliography{mybibfile}
\end{document}